\documentclass{article}
\usepackage{ijcai22}

\usepackage{times}
\usepackage{soul}
\usepackage{url}
\usepackage[hidelinks]{hyperref}
\usepackage[utf8]{inputenc}
\usepackage[small]{caption}
\usepackage{graphicx}
\usepackage{amsmath}
\usepackage{amsthm}
\usepackage{booktabs}
\usepackage{algorithm}
\usepackage{algorithmic}
\usepackage{xcolor}
\usepackage{amssymb}
\usepackage{enumerate}
\usepackage{svg}

\svgsetup{
    inkscapepath=i/svg-inkscape/
}
\svgpath{{svg/}}

\title{FastGCL: Fast Self-Supervised Learning on Graphs via \\Contrastive Neighborhood Aggregation}

\author{
Yuansheng Wang$^1$\and
Wangbin Sun$^1$\and
Kun Xu$^1$\and
Zulun Zhu$^2$\and
Liang Chen$^1$*\and
Zibin Zheng$^1$
\affiliations
$^1$Sun Yat-sen University\\
$^2$Nanyang Technological University
\emails
\{wangysh29, sunwb7, xukun6\}@mail2.sysu.edu.cn,
zulun001@ntu.edu.sg,\\
\{chenliang6, zhzibin\}@mail.sysu.edu.cn,
}

\begin{document}

\maketitle

\begin{abstract}
Graph contrastive learning (GCL), as a popular approach to graph self-supervised learning, has recently achieved a non-negligible effect.
To achieve superior performance, the majority of existing GCL methods elaborate on graph data augmentation to construct appropriate contrastive pairs.
However, existing methods place more emphasis on the complex graph data augmentation which requires extra time overhead, and pay less attention to developing contrastive schemes specific to encoder characteristics.
We argue that a better contrastive scheme should be tailored to the characteristics of graph neural networks (e.g., neighborhood aggregation) and propose a simple yet effective method named FastGCL.
Specifically, by constructing weighted-aggregated and non-aggregated neighborhood information as positive and negative samples respectively, FastGCL identifies the potential semantic information of data without disturbing the graph topology and node attributes, resulting in faster training and convergence speeds.
Extensive experiments have been conducted on node classification and graph classification tasks, showing that FastGCL has competitive classification performance and significant training speedup compared to existing state-of-the-art methods.
\end{abstract}

\section{Introduction}


Graph self-supervised learning (GSSL), which is capable of learning representations of graph structured data without artificially crafted labels, has numerous application scenarios in real systems \cite{survey2109}.
Recently, graph contrastive learning (GCL) \cite{DGI,GRACE,GraphCL} has made a big splash on GSSL, successfully applying contrastive learning from the field of computer vision to graph data.
By pulling the relevant pairs (\textit{positive}) closer and pushing the irrelevant pairs (\textit{negative}) farther on the latent space,  GCL methods achieve results comparable to, and sometimes surpassing, supervised methods.

One of the critical stages in contrastive learning is data augmentation, which aims to provide different views for contrastive pairs construction. 
Unlike images, the augmentation of graph data has to take not only the semantic but also the graph structural information into account. 
Existing GCL methods utilize feature shuffling \cite{DGI},  edge addition and removal \cite{GraphCL}, subgraph sampling \cite{GCC}, etc., for graph data augmentation.
The effectiveness of existing methods relies to their carefully designed data augmentation strategies \cite{AFGRL}.

However, these graph augmentation approaches mentioned above pose some inevitable problems. 
For one thing, the time consumption is increased when applying a complex augmentation algorithm.
For another thing, it disturbs the intrinsic structure of the original graph to a certain extent, accompanied with the risk of \textit{semantic drift} \cite{MoCL}, and it is difficult to determine whether the augmented graph is positively correlated with the original graph \cite{AFGRL}.
The majority of GCL methods follow the InfoMax \cite{InfoMax} principle, and aim to maximize mutual information of the corresponding graphs (or nodes) under different views of augmentation.
To correct the potential semantic drift introduced by these graph data augmentations, encoders need to learn an extra nonlinear transformation to rebuild the mapping between the augmented graph and the original graph, which leads to the slow convergence of training.

\begin{figure}[t]
\centering
  \includegraphics[width=\linewidth]{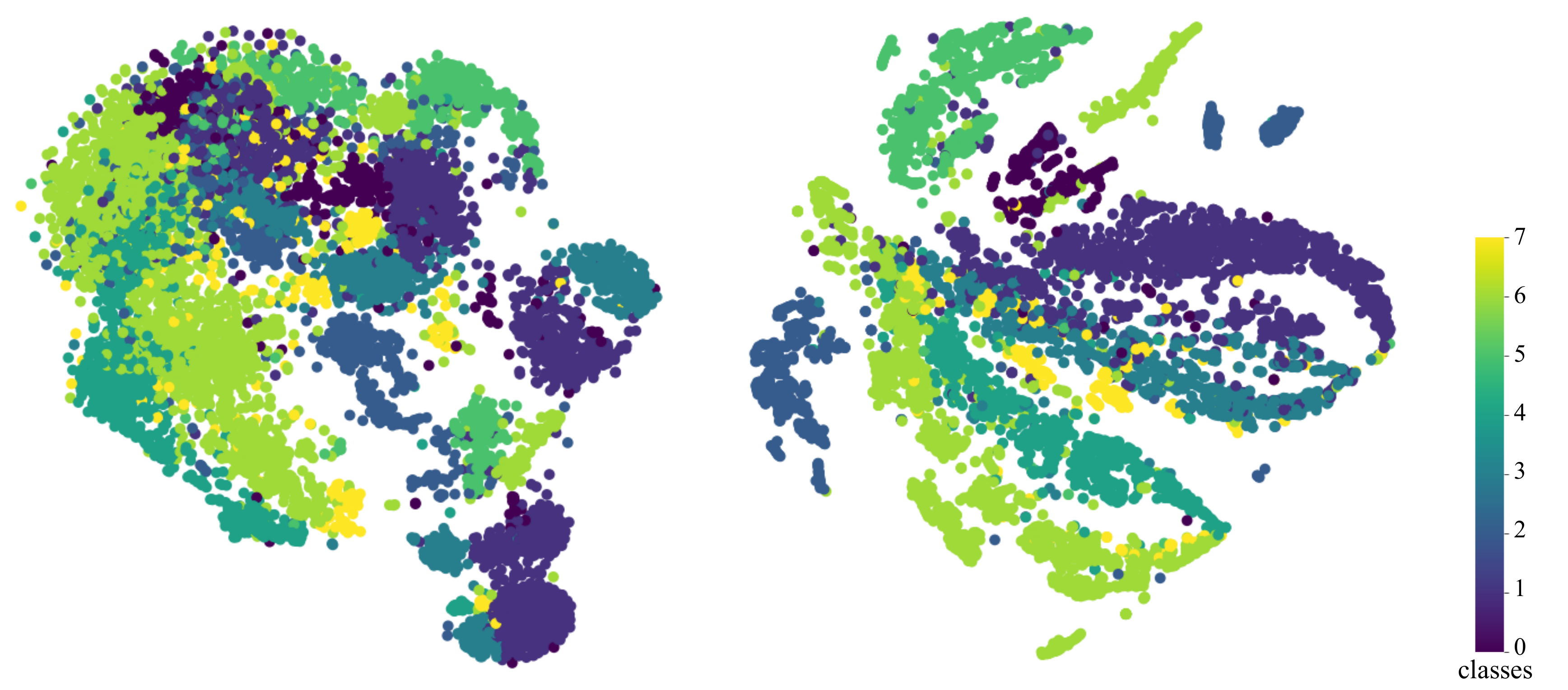}
\caption{t-SNE embeddings of nodes in the \textit{Amazon-Photo} dataset. 
A randomly initialized untrained SGC model can achieve discriminative representations (\textbf{right}) relying only on the neighbor aggregation, compared to raw features (\textbf{left}). }
\label{fig1}
\end{figure}

The limitations of existing augmentation strategies for GCL inspire us to look back to the characteristics of the GNN encoder itself.
In Figure \ref{fig1} we visualize the raw node features and the embedding of the randomly initialized untrained SGC \cite{SGC} activation layer.
It can be observed that without the help of trained model parameters, a more discriminative graph representation carrying semantics information can be obtained by neighbor information aggregation alone.
Thus, we argue that a better contrastive scheme should be tailored to the characteristics of GNNs (e.g., neighborhood aggregation).

Therefore, we propose a new solution for GCL, named \textbf{Fast} \textbf{G}raph \textbf{C}ontrastive \textbf{L}earnig (\textbf{FastGCL}).
Unlike previous methods, FastGCL utilizes contrastive neighborhood aggregation for self-supervised learning, without disturbing graph topology or node attributes.
Specifically, FastGCL generates different representations by using a shared GNN encoder. 
To obtain the positive, it weights each edge of the original graph for weighted neighborhood aggregation. 
Edge weights are computed by learning interactions between node representations of original graph output by the encoder. 
To obtain the negative, it simply choose the graph with only self-loop as input of the encoder. 
Our approach alleviates the time overhead of complex data augmentation strategies, as well as the potential risk of semantic drift from structural changes in graph data, which contributes to faster training and convergence speed compared to previous methods.

Our contributions can be summarized as follows:
\begin{itemize}
\item
We demonstrate the effectiveness of the GNN neighborhood aggregation in the unsupervised learning and suggest that a better graph contrastive scheme should be tailored to the characteristics of GNNs.
\item
We propose FastGCL, a fast and effective graph contrastive method in which the contrastive neighborhood aggregation is employed without disturbing the graph topology or node attributes.
\item
Comprehensive experiments on diverse datasets demonstrate that FastGCL is able to achieve competitive results with a remarkable training speedup compared to the state-of-the-art GCL methods.
\end{itemize}

\section{Related Work}\label{section2}

\begin{figure*}[t]
  \centering
  \includegraphics[width=\linewidth]{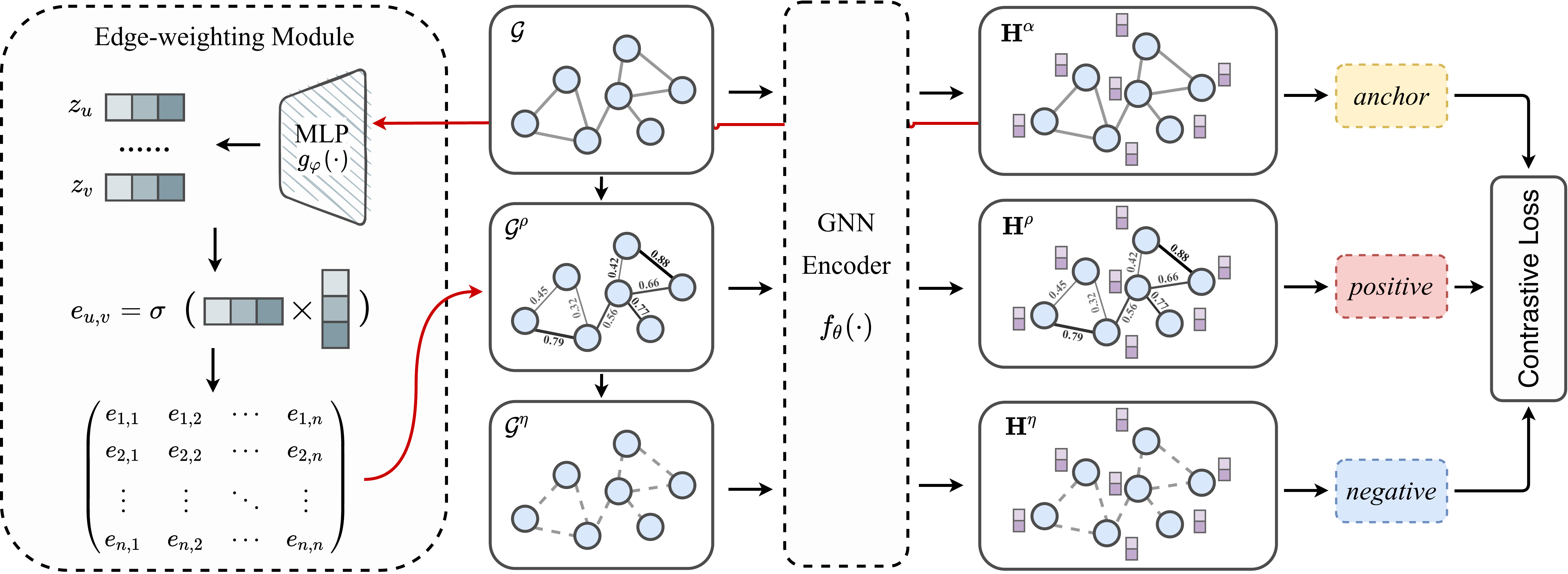}
  \caption{Overview of FastGCL. It mainly consists of a shared GNN encoder and a learnable edge-weighting module, and the latter takes the node representation $\mathbf{H}^\alpha$ of the original graph $\mathcal{G}$ as input (\textit{Red line}). The positive graph $\mathcal{G}^\rho$ is generated by the edge-weighting module, while the $\mathcal{G}^\eta$ with only self-loop (mathematically, $\mathbf{A}=\mathbf{I}$) is straightforwardly considered as the negative graph. All learnable parameters are optimized via contrastive loss.}
  \label{fig2}
\end{figure*}

Inspired by the field of computer vision, GCL methods have been explored for self-supervised learning on graphs with remarkable success.
The most substantial challenge for GCL is how to develop contrastive pairs on abstract and irregular graph data.
The general approach is performing data augmentation to generate different graph views.
If the augmented graphs (or nodes) originate from the same graph (or node), they are constructed as positive sample pairs (semantic correlation), otherwise they are negative sample pairs.
According to the strategies employed in current studies, we can categorize them into the following three groups:

\paragraph{Uniform Graph Data Augmentation.}
This strategy focuses on the structure and attributes of graph data and implements a non-discriminatory augmentation for all study objects (i.e., graphs or nodes).
For example, DGI \cite{DGI} generates negative graphs by corruption functions (e.g., random shuffling of features in row-wise).
GRACE \cite{GRACE} facilitates the optimization of contrastive objectives from two augmentation views through random edge removal and feature masking.
GCC \cite{GCC} uses ego-net sampling to contrst the L-hop neighbor subgraph of nodes.
GraphCL \cite{GraphCL} parameterizes four graph augmentation strategies (i.e., node dropping, edge perturbation, attribute masking, and subgraph sampling) to select best combination of augmentation.

\paragraph{Guided Graph Data Augmentation.}
Uniform graph data augmentation may perturb important nodes or edges and then affect the positive correlation of the same object in different views.
Therefore, some guided approaches emerge to identify the importance of nodes or edges through domain knowledge or learnable strategies to keep the semantics inherent in the augmented views.
GCA \cite{GCA} proposes adaptive data augmentation, specifically highlighting the important links based on node centrality, and adding noise to unimportant nodes.
MoCL \cite{MoCL} combines domain knowledge for molecular graph augmentation, without altering the molecular properties too much.
AD-GCL \cite{AD-GCL}, which extends the graph information bottleneck \cite{GIB} to unsupervised tasks, guides the process of graph data augmentation by a learnable perturbation model for edge removals.

\paragraph{No Negative Sampling or Augmentation.}
Instead of using negative sampling or data augmentation, some works looked for other information to construct contrastive pairs, which inspired our work to consider the characteristics of the encoder.
GMI \cite{GMI}, InfoGraph \cite{InfoGraph} do not require explicit data augmentation, the former maximizes the mutual information of edge and node features on the original graph, and the latter constructs contrastive pairs in a batch-wise fashion.
BGRL \cite{BGRL} is based on the BYOL \cite{BYOL} framework without the use of negative samples, which avoids the sampling bias \cite{samplebias}.
AFGRL \cite{AFGRL} discovers nodes that can be treated as positive samples by considering the local structural information and global semantics of the graph, without creating an augmented graph view and negative sampling.

Most of the aforementioned methods, however, disturb the structural and attribute information when providing different views for graphs.
In contrast, our approach tends to preserve topology and attributes. 
We use a learnable edge weighting module to provide the positive graph view and preserve semantic information to the maximum extent.
The design for the negative sample is straightforwardly based on GNNs' aggregation mechanism, which avoids a large amount of negative samples and sampling bias.
Thus, by considering the difference of neighbor information aggregation as a contrastive objective, our augmentation strategy is more suitable for GNNs.

\section{Proposed Method}

\subsection{Notations and Preliminaries}
Let $ \mathcal{G} = (\mathcal{V}, \mathcal{E}) $ denote an undirected graph, where $ \mathcal{V} = \{v_1, v_2, \cdots, v_n\} $ represents all nodes, $ N = | \mathcal{V} | $, and $ \mathcal{E} $ is the set of all edges, $ M = | \mathcal{E} | $.
The matrix $ \mathbf{X} \in \mathbb{R} ^ {N \times F} $ is composed of $ F $-dimensional attributes $ x_v $ of node $ v $ on graph $ \mathcal{G} $. 
The symmetric matrix $ \mathbf{A} \in \mathbb{R} ^ {N \times N} $ denotes the adjacency matrix where $A_{i j}=1$, if $(v_i,v_j) \in \mathcal{E} $, otherwise $ A_{i j}=0 $. 
In addition, we use the matrix $ \mathbf{E} \in \mathbb{R} ^ {N \times N} $ to denote the weights on the corresponding edges of the adjacency matrix, which are floating point numbers between 0 and 1. 
$ \mathcal{N}_v $ is the set of neighbors of the node $ v $ and $ \mathbf{D} $ is the degree matrix $ {D}_{i i} = \sum^{N}_{j=0} A_{i j}$. 
We denote the symmetric renormalized adjacency matrix $ \mathbf{\hat{A}} $ by $ \mathbf{\hat{A}} = \mathbf{\tilde{D}^{1/2}\tilde{A}\tilde{D}^{1/2}} $, where $ \mathbf{\tilde{A}} = \mathbf{A} + \mathbf{I} $ is the adjacency matrix with self-loop and $ \mathbf{I} $ is the identity matrix.

In our work, we use GNN as our encoder. It typically uses a recursive neighborhood aggregation protocol where the representation $ \mathbf{h}_v $ of node $ v $ is obtained by aggregating its neighborhood representations by $ agg(\cdot)$ at each iteration, which is then updated by a nonlinear function $ upd(\cdot) $. The representation of a node can capture the semantic and structural information of its $ k $-hop neighbors after $ k^{th} $ iterations of aggregation. The process can be expressed as,
\begin{equation}\label{eq1}
   \mathbf{h}_{v}^{(k)}=\textit{upd}^{(k)}\left(\mathbf{h}_{v}^{(k-1)}, \textit{ agg}^{(k)}\left(\left\{\mathbf{h}_{u}^{(k-1)} \mid u \in \mathcal{N}_{v}\right\}\right)\right)
\end{equation}
where the $ \mathbf{x}_v $ is as the initialization $ \mathbf{h}_v^{(0)}$ of the node representation.

Our goal is to train an encoder  $ f_\theta(\cdot) $, given a graph $ \mathcal{G} $ along with $ \mathbf{X} $ and $ \mathbf{A} $, to learn a high-level node representation $ \mathbf{H} = f_\theta(\mathbf{A}, \mathbf{X})$ which can be generalized to various downstream tasks, without using any label information of categories.

\subsection{FastGCL}\label{section 3.2}
The overall framework of FastGCL is shown in Figure \ref{fig2}.
The generation of graph views and the objective function are two vital components of GCL methods, which are introduced in this section.

\paragraph{Graph View Generation.}
Standard contrastive learning approaches implicitly assume that the positive and the negative belong to different categories in downstream tasks \cite{SaunshiPAKK19}. 
\cite{GCA} states that data augmentation strategies should preserve the intrinsic structure and attributes of the graph. 
Our design adheres to the above constraints.
Considering the representation of original graph $\mathcal{G}$ as the anchor, we need to construct the positive and the negative under two other graph views (i.e., $\mathcal{G}^\rho$ and $\mathcal{G}^\eta$) for contrastive learning.
This process can be described as follows.

Firstly, with a GNN encoder $f_\theta(\cdot)$ , all node representations of the original graph $\mathbf{H}^\alpha$ can be obtained. Mathematically,
\begin{equation}\label{eq2}
\mathbf{H}^\alpha = f_\theta(\mathbf{A}, \mathbf{X})
\end{equation}
$\mathbf{H}^\alpha$ aggregates rich information about each node and its neighbors. 
We take $\mathbf{H}^\alpha$ as input of an edge-weighting module to generate the weights of the corresponding edges.
This process involves the help of a MLP $g_\varphi(\cdot)$, which projects the node representations to a new latent space, denoted as,
\begin{equation}\label{eq3}
\mathbf{Z} = g_\varphi(\mathbf{H}^\alpha)
\end{equation}
\begin{equation}\label{eq4}
e_{u,v} = \left\{
\begin{matrix}
 \sigma(\mathbf{z}_u \cdot \mathbf{z}_v^{\top}), & \textit{ if }(u,v) \in \mathcal{E}
\\  \\
 0,  & \textit{ otherwise. }
\end{matrix}
\right.
\end{equation}
where $\sigma$ is sigmoid function and $e_{u,v}$ represents the weight between node $u$ and node $v$,  if there is an edge between them.
Then these weights are performed a element-wise product with the adjacency matrix $\mathbf{A}$ of the original graph. 
The positive representation $\mathbf{H}^\rho$ is then obtained by the shared GNN encoder $f_\theta(\cdot)$, represented as,
\begin{equation}\label{eq5}
\mathbf{H}^\rho = f_\theta(\mathbf{A} \odot \mathbf{E}, \mathbf{X})
\end{equation}
This is equivalent to the fact that we bring these learned edge weights to the GNN for weigted neighborhood aggregation, instead of equal aggregation for Equation \ref{eq2}. 
For the negative, we simply choose the identity matrix $\mathbf{I}$ as the shared GNN encoder's input,
\begin{equation}\label{eq6}
\mathbf{H}^\eta  = f_\theta(\mathbf{I}, \mathbf{X})
\end{equation}
By taking the identity matrix $\mathbf{I}$ as input, the GNN will degenerate into an MLP, which can only aggregate information from the nodes themselves.
It is suitable to serve as an uncorrelated negative representation.

With this design, for the positive sample, it does not lose neighbor information during aggregation, but only affects how much or how little neighbor information each node receives.
This weighted aggregation of neighbors brings a difference in representation that can be seen as an augmentation, allowing the encoder to better extract positively relevant features.
For the negative sample, it loses all information about its neighbors, in which case it is likely to be predicted as another category by the downstream task.
As we know that the information of the graph data is mainly determined by the composition of the node and its neighbors, see Equation \ref{eq1}.

\paragraph{Objective Function.}
Inspired by noise-contrastive estimation (NCE) \cite{NCE}, to pull the positive closer and push the negative  farther, our self-supervised learning loss is designed as,
\begin{equation}\label{eq7}
    \mathcal{L}_{\mathrm{ssl}}=-\frac{1}{N} \sum_{v \in \mathcal{V}} \left[\log\sigma(sim^\rho_v) + \log(1-\sigma(sim^\eta_v)) \right]
\end{equation}
where $sim$ is the similarity score between two representations, and the positive and negative samples similarity scores of node $v$ can be expressed as,
\begin{equation}\label{eq8}
    sim^\rho_v = \ell(\mathbf{h}_v^\alpha, \mathbf{h}_v^\rho), \, \,
    sim^\eta_v = \ell(\mathbf{h}_v^\alpha, \mathbf{h}_v^\eta)
\end{equation}
where $\ell$ is the cosine similarity function.
For the node classification task, our contrastive objective is the corresponding nodes in different views.
For the graph classification task, our aim is to learn discriminative and semantic graph representations, so our contrastive objective becomes the graph representations of different views.
The representation of graph $\mathcal{G}$, $\mathbf{h}_\mathcal{G}$, can be obtained by a pooling operation that combines the representations of all nodes in $K$ iterations. It is formally defined as,
\begin{equation}\label{eq9}
   \mathbf{h}_\mathcal{G} = \textit{pool}\left(\left\{\mathbf{h}_{v}^{(K)} \mid v \in \mathcal{V}\right\}\right)
\end{equation}

To prevent the positive graph from becoming almost identical to the original graph, which would lead to an indistinguishable representation compared with the anchor, a regularization term is applied. Mathematically,
\begin{equation}\label{eq10}
    \mathcal{L}_{\mathrm{norm}}= -\frac{1}{M} \sum_{i=1}^{N}\sum_{j=1}^{N} \log\left(1 + \exp\left(\mathbf{A}-\mathbf{E}\right)\right)
\end{equation}
which makes the $\mathbf{E}$ less close to the $\mathbf{A}$ to a better learning. In conjunction with Equation \ref{eq7}, we propose an objective function that is to minimize the loss $\mathcal{L}$ of the entire network, i.e.,
\begin{equation}\label{eq11}
    \mathrm{\mathop{min}\limits_{\theta, \varphi}} \, \mathcal{L}=  \mathcal{L}_{\mathrm{ssl}} + \lambda\mathcal{L}_{\mathrm{norm}}
\end{equation}
where $\lambda$ is a hyper-parameter to regulate the strength of the penalty.

In summary, \textbf{1)} FastGCL constructs contrastive pairs for the shared GNN encoder without disturbing the topology and attributes of the graph and learns the representation by contrasting the neighborhood aggregations. 
\textbf{2)} The positive samples of FastGCL are obtained by a learnable edge-weighting module, which can guide the weighted graph to maintain a certain amount of information associated with downstream tasks. 
The negative examples, on the other hand, are straightforward in design and do not require the involvement of complex augmentation functions.

\section{Experiments}

\begin{table*}[t]
\centering
\small
\caption{Node classification performance (Accuracy). \textit{R.I.U.} indicates randomly initialized untrained model.}
\begin{tabular}{lcccccc}
\toprule
  \textbf{Method} &
  \textbf{WikiCS} &
  \textbf{Computers} &
  \textbf{Photo} &
  \textbf{Co.CS} &
  \textbf{Co.Physics} \\ \midrule
GCN \cite{GCN}      & 78.85 ± 0.09 & 86.42 ± 0.09 & 92.71 ± 0.03 & 93.07 ± 0.02 & 95.90 ± 0.01  \\
GAT \cite{GAT}      & 78.93 ± 0.05 & 88.30 ± 0.04 & 92.10 ± 0.04 & 92.71 ± 0.04 & 95.91 ± 0.01  \\
SGC \cite{SGC}      & 78.63 ± 0.05 & 87.36 ± 0.04 & 92.95 ± 0.04 & 93.03 ± 0.02 & \text{OOM} \\ \midrule
GCN \textit{R.I.U.} & 77.97 ± 0.09 & 88.16 ± 0.12 & 92.72 ± 0.07 & 93.14 ± 0.04 & 95.72 ± 0.03 \\
GAT \textit{R.I.U.} & 78.03 ± 0.05 & 85.72 ± 1.21 & 91.69 ± 0.38 & 91.85 ± 0.14 & 95.21 ± 0.04 \\
SGC \textit{R.I.U.} & 77.44 ± 0.10 & 88.30 ± 0.13 & 92.83 ± 0.07 & 93.17 ± 0.03 & 95.71 ± 0.03\\
Raw feature \cite{GCA} & 71.98 ± 0.00 & 73.81 ± 0.00 & 78.53 ± 0.00 & 90.37 ± 0.00 & 93.58 ± 0.00 \\\midrule
DGI \cite{DGI}      & 77.31 ± 0.10 & 87.09 ± 0.14 & 92.19 ± 0.12 & 92.68 ± 0.06 & \text{OOM}    \\
GRACE \cite{GRACE}  & 78.50 ± 0.19 & 89.53 ± 0.35 & 92.78 ± 0.45 & 91.12 ± 0.02 & \text{OOM}    \\
GCA \cite{GCA}      & 78.35 ± 0.05 & 87.85 ± 0.31 & 92.53 ± 0.16 & 93.10 ± 0.01 & 95.73 ± 0.03  \\
BGRL \cite{BGRL}    & \textbf{79.98} ± \textbf{0.10} & \textbf{90.34} ± \textbf{0.19} & 93.17 ± 0.30 & \textbf{93.31} ± \textbf{0.13} & \textbf{95.73} ± \textbf{0.05} \\
AFGRL \cite{AFGRL}  & 77.62 ± 0.49 & 89.88 ± 0.33 & \textbf{93.22} ± \textbf{0.28} & 93.27 ± 0.17 & 95.69 ± 0.10 \\ \midrule
\textbf{FastGCL} (Ours)    & 79.20 ± 0.07  & 89.35 ± 0.09 & 92.91 ± 0.07 & 92.71 ± 0.07 & 95.53 ± 0.02 \\ \bottomrule
\end{tabular}
\label{tab1}
\end{table*}

\begin{table*}[t]
\centering
\small
\caption{Graph classification performance (Accuracy).}
\begin{tabular}{lcccccc}
\toprule
  \textbf{Method} &
  \textbf{NCI1} &
  \textbf{PROTEINS} &
  \textbf{DD} &
  \textbf{COLLAB} &
  \textbf{RDT-B} &
  \textbf{RDT-M5K} \\ \midrule
  GIN \cite{GIN}              & 76.37 ± 1.93 & 73.69 ± 1.79 & 74.53 ± 1.57 & 73.62 ± 0.99 & 87.35 ± 2.61 & 53.29 ± 1.14 \\
  GIN \textit{R.I.U.}         & 62.35 ± 1.00 & 63.67 ± 0.67 & 60.00 ± 1.37 & 63.61 ± 1.28 & 63.12 ± 1.34 & 39.23 ± 1.13 \\ \midrule
  InfoGraph \cite{InfoGraph}  & 69.42 ± 0.58 & 73.03 ± 0.26 & 69.57 ± 0.95 & 68.78 ± 0.47 & 75.58 ± 0.71 & 45.59 ± 0.84 \\
  GCC \cite{GCC}              & 62.41 ± 0.13 & 72.61 ± 0.23 & 74.09 ± 0.15 & \textbf{71.16} ± \textbf{0.04} & 77.97 ± 0.24 & 49.31 ± 0.20 \\
  GraphCL \cite{GraphCL}      & 70.89 ± 1.25 & 73.02 ± 0.51 & 70.00 ± 1.17 & 69.40 ± 0.72 & 76.11 ± 2.38 & 46.73 ± 0.39 \\
  AD-GCL \cite{AD-GCL}        & 71.74 ± 0.42 & 73.16 ± 0.51 & 75.00 ± 0.47 & 69.85 ± 0.28 & 78.49 ± 1.04 & 46.36 ± 0.78 \\ \midrule
  \textbf{FastGCL} &
    \textbf{72.68} ± \textbf{0.70} &
    \textbf{75.77} ± \textbf{0.28} &
    \textbf{76.61} ± \textbf{0.44} &
    69.34 ± 0.40 &
    \textbf{82.75} ± \textbf{2.70} &
    \textbf{51.50} ± \textbf{0.91} \\ \bottomrule
\end{tabular}
\label{tab2}
\end{table*}

\subsection{Experimental Settings}

\paragraph{Datasets.}
To evaluate the performance of our proposed method on different graph tasks, we conduct experiments on diverse domain datasets. 
The performance of node classification is evaluated on 5 medium-scale and widely-used datasets \cite{wikics,pitfalls}. 
The performance of graph classification is evaluated on 6 datasets with or without node features, involving molecules, proteins and social networks \cite{tudataset}. 

\paragraph{Evaluation Protocol.}
We train all models in an unsupervised manner. 
For node classification, we use a standard linear evaluation protocol \cite{DGI} to evaluate the quality of the embeddings learned by the model. 
Specifically, the parameters and gradients of the model are frozen and the embeddings are used as input for classification using a logistic regression with the $\ell_2$ regularization. 
In addition to using the public 20 divisions on the WikiCS dataset, we randomly divide the training set, validation set, and test set 20 times according to 10\%, 10\%, and 80\% on the other datasets \cite{BGRL}, and report the best round on average performance of the validation set as the evaluation result. 
For graph classification, \cite{AD-GCL} specifically states that downstream classifiers can significantly affect the evaluation results of unsupervised learning. 
We therefore uniformly use the standard linear evaluation protocol instead of SVM to evaluate all models and use ten-fold cross-validation to evaluate the performance of graph classification.

\paragraph{Implementation Details.}
We use GCN \cite{GCN} with edge weights as an encoder for the node-classification task, and the node-wise expression is as follows.
\begin{equation}\label{eq12}
    \mathbf{h}_{v}^{(k)}=\boldsymbol{\Theta}^{\top} \sum_{u \in \mathcal{N}_v \cup\{v\}} \frac{e_{u, v}}{\sqrt{\hat{d}_{u} \hat{d}_{v}}} \mathbf{h}_{u}^{(k-1)}
\end{equation}
GIN \cite{GIN} with edge weights is applied as an encoder for the graph classification task. It can be formally denoted as,
\begin{equation}\label{eq13}
    \mathbf{h}_{v}^{(k)}=\textit{MLP}_{\Theta}\left((1+\epsilon) \cdot \mathbf{h}_{v}^{(k-1)} + \sum_{u \in \mathcal{N}_v}  e_{u, v} \cdot \mathbf{h}_{u}^{(k-1)}\right)
\end{equation}
The sum pooling is the readout function for graph representation and learn layer-wise graph representation, 
\begin{equation}\label{eq14}
    \mathbf{h}_{\mathcal{G}}=\mathop{\parallel}\limits_{k=1}^{K} \sum_{v \in \mathcal{V}} \mathbf{h}_v^{(k)}
\end{equation}
where $\parallel$ denotes the concatenation operation and $K$ is the number of layers.
For graph data without node features, we simply make a 1-dimensional unit vector as node features. 
These settings above are kept consistent with all baseline models.
We implement our proposed method based on the PyTorch and PyG\footnote{https://www.pyg.org/} code libraries and evaluate embeddings more quickly using a logistic regression classifier on CUDA.
The Adam algorithm with weight decay was used as the optimizer for our model. 
Hyperparameters such as learning rate, regular strength $\lambda$, embedding dimension size, and number of layers of the model were fully examined with grid search.

\paragraph{Baselines.}
We compare the representative GCL methods mentioned in Section \ref{section2}. 
Since the encoders of the existing methods are basically GCN or GIN, the difference between their encoders may only be the number of layers, dimension size, regularization, and implementation. 
Therefore, we additionally evaluate the performance of randomly initialized untrained GNNs for node classification (i.e., GCN, GAT, SGC) and graph classification (i.e., GIN), respectively, to explore the effect brought by GNN neighbor aggregation. 
Notably, since our evaluation using logistic regression on CUDA would be a little worse than the previously officially reported results for the graph classification task using SVM, we re-report our experimental results. 
On the node classification task, there is not much difference with the previous publicly available results, so here we directly report them, except for DGI (could be better) and GRACE on \textit{WikiCS} (slightly worse).
Here we report our reproduced results of them.
All our experiments were run ten times to avoid bias from random initialization under the same random division.

\subsection{Experimental Analysis}
\paragraph{Overall Performance.}
The experimental results of node classification and graph classification are summarized in Table \ref{tab1} and Table \ref{tab2}, respectively.
We can observe from them that:
\textbf{1)} Our proposed method FastGCL achieves competitive classification performance on node classification compared to recent GCL method and outperforms supervised GNNs on some datasets (e.g., \textit{WikiCS}, \textit{Computers}).
For the graph classification task, our model exceeds the state-of-the-art methods on most of datasets. 
There is a significant improvement on datasets without node features (i.e., \textit{RDT-B}, \textit{RDT-M5K}), which is closer to the performance of supervised GIN compared to previous methods.
\textbf{2)} Randomly initialized untrained GNNs also has relatively good performances, even better than DGI, a pioneering work of GCL, on the node classification task. In comparison with the evaluation performances of raw feature, the node representation obtained by neighbor aggregation of GNNs helps the downstream classifier achieve better performance, especially on \textit{Computers}.
\textbf{3)} Our method improves on the basis of randomly initialized untrained GNNs, indicating that our contrastive scheme is effective. 
This improvement is obvious on the \textit{WikiCS} that contains dense node features.
Moreover, on the graph classification datasets without rich node features, our contrastive scheme drives GNN encoder to learn richer potential information from the graph structure, obtaining a significant improvement in classification performance compared to randomly initialized untrained GIN.

\begin{figure}[ht]
\centering
\includegraphics[width=\linewidth]{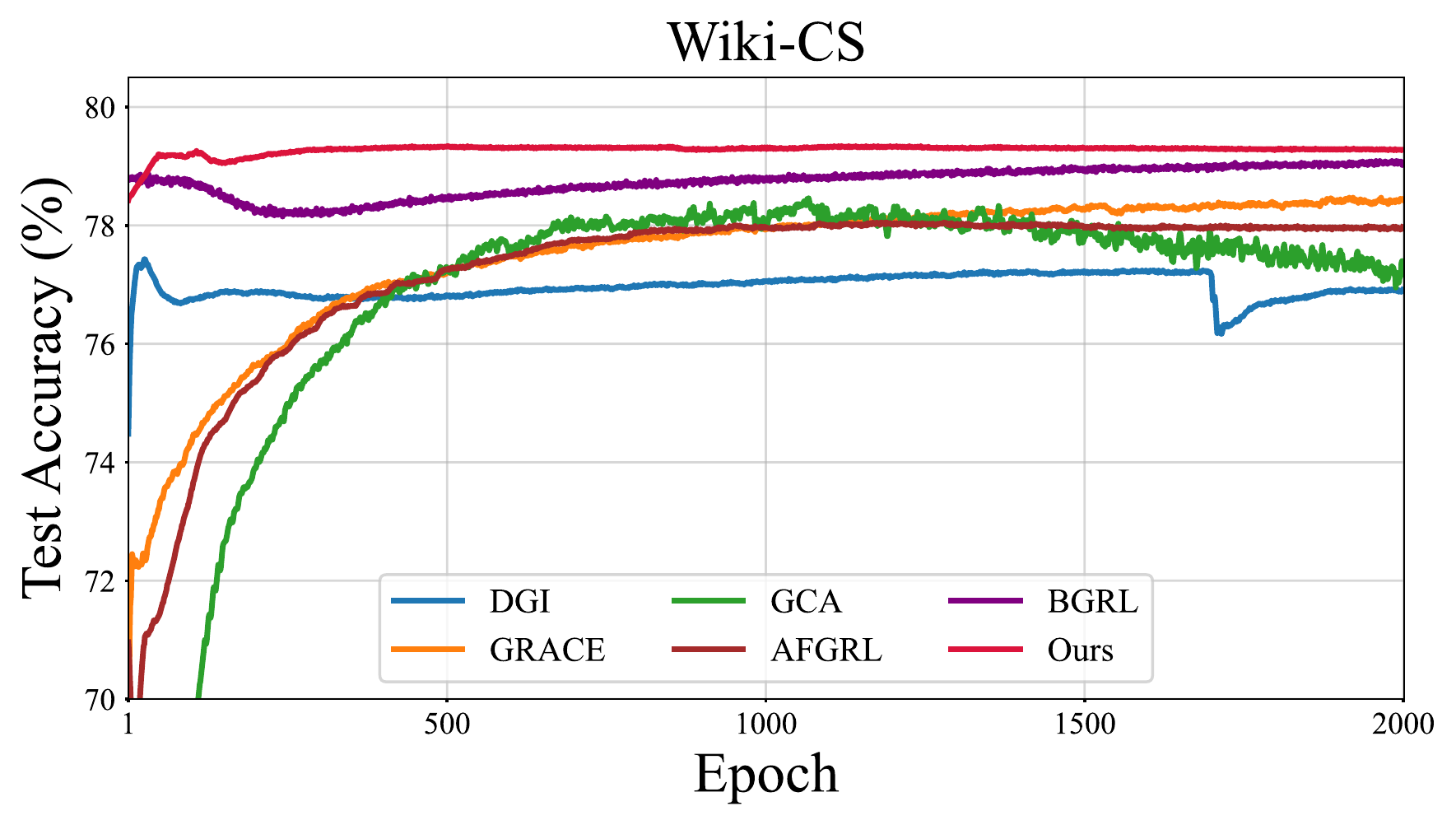}
\caption{Learning curve of all node classification baselines on \textit{WikiCS}.}
\label{fig3}
\end{figure}

\paragraph{Training Efficiency.}
In Figure \ref{fig3}, we plot the learning curve of all models on the test set of \textit{WikiCS} on the node classification task.
It can be noticed that FastGCL converges quickly, with less than 100 epochs. While other models need longer training epochs for unsupervised learning.
In particular, BGRL, which uses the BYOL framework, maintained its upward trend after 300 epochs, but does not obtain its best results until training exceeded 9,000 epochs.
Within 2000 training epochs, FastGCL outperforms all the comparison methods.
The fast convergence of FastGCL is due to the absence of structure-changed perturbations, while other models need to implicitly learn a nonlinear mapping to correct the potential semantic drift brought by data augmentation. Besides, it benefits from the contrastive learning designed for GNN characteristics, contrasting weighted and unweighted neighborhood aggregation to quickly optimize the parameters of the GNN aggregation function.

\begin{figure}[t]
\includegraphics[width=\linewidth]{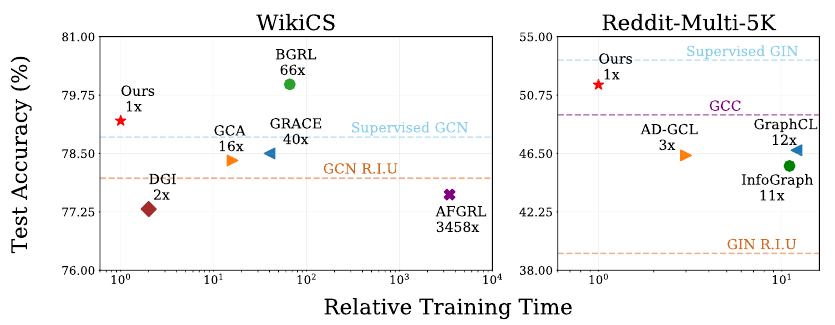}
\caption{Performance over training time on \textit{WikiCS} and \textit{RDT-M5K}.}
\label{fig4}
\end{figure}

Figure \ref{fig4} shows the training time performance of all models on WikiCS and RDT-M5K\footnote{Since GCC is a general pre-trained graph model that is not specific to a specific dataset, we have not compared it.}, respectively, relative to that of FastGCL. 
It can be noticed that FastGCL is faster than all previous methods while achieving competitive classification accuracy. 
The methods using complex data augmentation need more time to construct contrastive pairs.
Meanwhile, the method that does not use data augmentation, AF-GCL, is obviously more time consuming because of the clustering operation used to identify potential positive sample nodes.

\begin{figure}[ht]
\centering
\includegraphics[width=\linewidth]{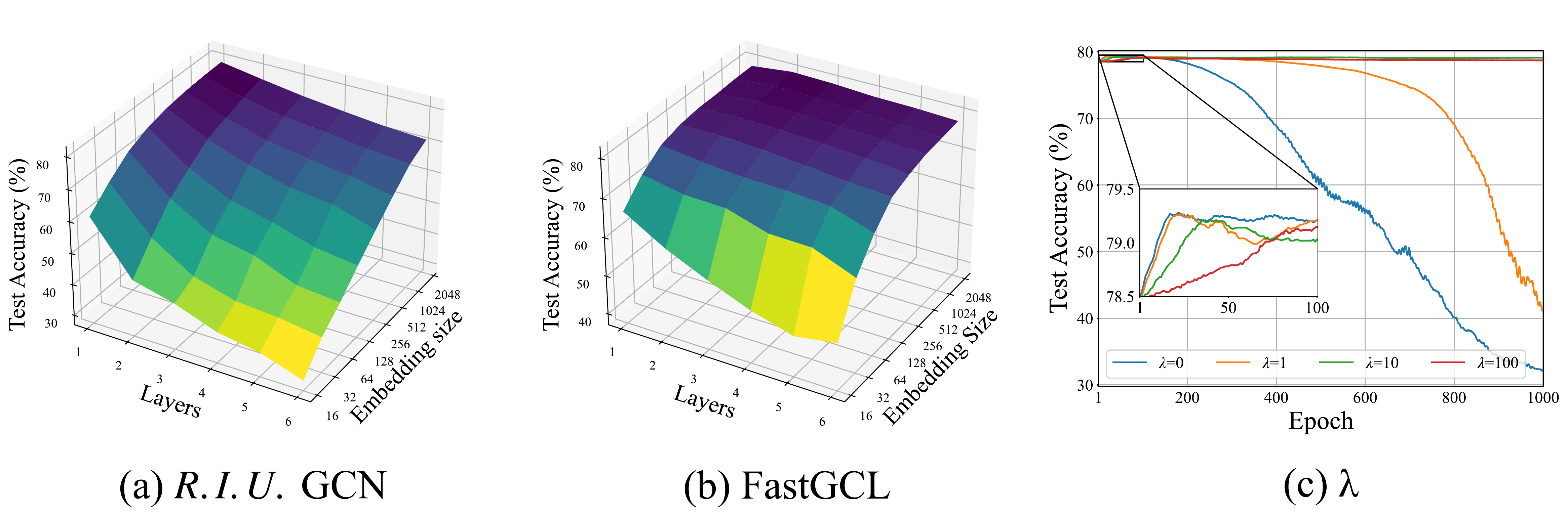}
  \caption{Effects of key parameters on \textit{WikiCS}.}
  \label{fig5}
\end{figure}

\paragraph{Sensitivity Analysis.}
We performed sensitivity analysis on several key parameters of FastGCL on \textit{WikiCS}, namely the embedding dimension size $d$ and the number of layers $l$ of GNN, as well as the regular strength $\lambda$.
The effects of $d$ and $l$ on the randomly initialized untrained GCN and FastGCL are demonstrated in Figure \ref{fig5} in (a) and (b), respectively.
As can be seen, the FastGCL trend is mostly consistent with the randomly initialized untrained GCN, and it can be inferred that the wider dimension and the shallower layers the GNN encoder has, the better it performs in the downstream task evaluation.
The Figure 5(c) demonstrates the effect of the size of $\lambda$ on the FastGCL learning stage. Too small $\lambda$ can cause the collapse of learning while too large $\lambda$ can lead to a slower convergence. A suitable $\lambda$ can make our contrastive learning more stable.

\paragraph{Ablation Study.}
An ablation study has been conducted to explore the effects of our edge-weighting method with the same hyper-parameters for fair comparison.
The results can find in Table \ref{tab3}. 
It can be concluded that the edge-weighting method we designed is effective in guiding the model to discriminate the importance of the graph topology and highlight critical connections compared to random weighting.

\begin{table}[ht]
\caption{Performance of different edge-weighting methods.}
\small
\centering
\begin{tabular}{lccc}
\toprule
\textbf{Datatset} & \textbf{Random Weighting} & \textbf{FastGCL}    \\ \midrule
WikiCS            & 79.11 ± 0.08 & \textbf{79.20 ± 0.07} \\
Computers         & 89.08 ± 0.09 & \textbf{89.35 ± 0.09} \\ \midrule
NCI1              & 68.18 ± 1.09 & \textbf{72.68 ± 0.70} \\
RDT-B             & 76.76 ± 3.24 & \textbf{82.75 ± 2.70} \\ \bottomrule
\end{tabular}
\label{tab3}
\end{table}



\section{Conclusion}
In this paper, we propose a simple yet effective GCL method termed FastGCL.
Specifically, it performs contrastive learning by contrasting weighted-aggregated and non-aggregated neighborhood information, rather than disturbing the topology and attributes of graph data to generate augmented views.
This aggregation-based contrastive scheme provides a new perspective for GCL research.
Extensive experiments demonstrate that FastGCL has an advantage in training time and can achieve competitive performance in downstream tasks.

\clearpage
\bibliographystyle{named}
\bibliography{ijcai22}

\begin{thebibliography}{}

\bibitem[\protect\citeauthoryear{Bielak \bgroup \em et al.\egroup
  }{2021}]{samplebias}
Piotr Bielak, Tomasz Kajdanowicz, and Nitesh~V. Chawla.
\newblock Graph barlow twins: A self-supervised representation learning
  framework for graphs.
\newblock {\em arXiv preprint arXiv: Arxiv-2106.02466}, 2021.

\bibitem[\protect\citeauthoryear{Grill \bgroup \em et al.\egroup }{2020}]{BYOL}
Jean{-}Bastien Grill, Florian Strub, Florent Altch{\'{e}}, Corentin Tallec,
  Pierre~H. Richemond, Elena Buchatskaya, Carl Doersch, Bernardo~{\'{A}}vila
  Pires, Zhaohan Guo, Mohammad~Gheshlaghi Azar, Bilal Piot, Koray Kavukcuoglu,
  R{\'{e}}mi Munos, and Michal Valko.
\newblock Bootstrap your own latent - {A} new approach to self-supervised
  learning.
\newblock In {\em NeurIPS}, 2020.

\bibitem[\protect\citeauthoryear{Gutmann and Hyv{\"{a}}rinen}{2010}]{NCE}
Michael Gutmann and Aapo Hyv{\"{a}}rinen.
\newblock Noise-contrastive estimation: {A} new estimation principle for
  unnormalized statistical models.
\newblock In {\em {AISTATS}}, volume~9 of {\em {JMLR} Proceedings}, pages
  297--304. JMLR.org, 2010.

\bibitem[\protect\citeauthoryear{Kipf and Welling}{2017}]{GCN}
Thomas~N. Kipf and Max Welling.
\newblock Semi-supervised classification with graph convolutional networks.
\newblock In {\em {ICLR} (Poster)}. OpenReview.net, 2017.

\bibitem[\protect\citeauthoryear{Lee \bgroup \em et al.\egroup }{2021}]{AFGRL}
Namkyeong Lee, Junseok Lee, and Chanyoung Park.
\newblock Augmentation-free self-supervised learning on graphs.
\newblock {\em arXiv preprint arXiv: Arxiv-2112.02472}, 2021.

\bibitem[\protect\citeauthoryear{Linsker}{1988}]{InfoMax}
Ralph Linsker.
\newblock An application of the principle of maximum information preservation
  to linear systems.
\newblock In {\em {NIPS}}, pages 186--194. Morgan Kaufmann, 1988.

\bibitem[\protect\citeauthoryear{Mernyei and Cangea}{2020}]{wikics}
Péter Mernyei and Cătălina Cangea.
\newblock Wiki-cs: A wikipedia-based benchmark for graph neural networks.
\newblock {\em arXiv preprint arXiv: Arxiv-2007.02901}, 2020.

\bibitem[\protect\citeauthoryear{Morris \bgroup \em et al.\egroup
  }{2020}]{tudataset}
Christopher Morris, Nils~M. Kriege, Franka Bause, Kristian Kersting, Petra
  Mutzel, and Marion Neumann.
\newblock Tudataset: A collection of benchmark datasets for learning with
  graphs.
\newblock {\em arXiv preprint arXiv: Arxiv-2007.08663}, 2020.

\bibitem[\protect\citeauthoryear{Peng \bgroup \em et al.\egroup }{2020}]{GMI}
Zhen Peng, Wenbing Huang, Minnan Luo, Qinghua Zheng, Yu~Rong, Tingyang Xu, and
  Junzhou Huang.
\newblock Graph representation learning via graphical mutual information
  maximization.
\newblock In {\em {WWW}}, pages 259--270. {ACM} / {IW3C2}, 2020.

\bibitem[\protect\citeauthoryear{Qiu \bgroup \em et al.\egroup }{2020}]{GCC}
Jiezhong Qiu, Qibin Chen, Yuxiao Dong, Jing Zhang, Hongxia Yang, Ming Ding,
  Kuansan Wang, and Jie Tang.
\newblock {GCC:} graph contrastive coding for graph neural network
  pre-training.
\newblock In {\em {KDD}}, pages 1150--1160. {ACM}, 2020.

\bibitem[\protect\citeauthoryear{Saunshi \bgroup \em et al.\egroup
  }{2019}]{SaunshiPAKK19}
Nikunj Saunshi, Orestis Plevrakis, Sanjeev Arora, Mikhail Khodak, and
  Hrishikesh Khandeparkar.
\newblock A theoretical analysis of contrastive unsupervised representation
  learning.
\newblock In {\em {ICML}}, volume~97 of {\em Proceedings of Machine Learning
  Research}, pages 5628--5637. {PMLR}, 2019.

\bibitem[\protect\citeauthoryear{Shchur \bgroup \em et al.\egroup
  }{2018}]{pitfalls}
Oleksandr Shchur, Maximilian Mumme, Aleksandar Bojchevski, and Stephan
  Günnemann.
\newblock Pitfalls of graph neural network evaluation.
\newblock {\em arXiv preprint arXiv: Arxiv-1811.05868}, 2018.

\bibitem[\protect\citeauthoryear{Sun \bgroup \em et al.\egroup
  }{2019}]{InfoGraph}
Fan-Yun Sun, Jordan Hoffmann, Vikas Verma, and Jian Tang.
\newblock Infograph: Unsupervised and semi-supervised graph-level
  representation learning via mutual information maximization.
\newblock {\em arXiv preprint arXiv: Arxiv-1908.01000}, 2019.

\bibitem[\protect\citeauthoryear{Sun \bgroup \em et al.\egroup }{2021}]{MoCL}
Mengying Sun, Jing Xing, Huijun Wang, Bin Chen, and Jiayu Zhou.
\newblock Mocl: Contrastive learning on molecular graphs with multi-level
  domain knowledge.
\newblock {\em arXiv preprint arXiv: Arxiv-2106.04509}, 2021.

\bibitem[\protect\citeauthoryear{Suresh \bgroup \em et al.\egroup
  }{2021}]{AD-GCL}
Susheel Suresh, Pan Li, Cong Hao, and Jennifer Neville.
\newblock Adversarial graph augmentation to improve graph contrastive learning.
\newblock {\em arXiv preprint arXiv: Arxiv-2106.05819}, 2021.

\bibitem[\protect\citeauthoryear{Thakoor \bgroup \em et al.\egroup
  }{2021}]{BGRL}
Shantanu Thakoor, Corentin Tallec, Mohammad~Gheshlaghi Azar, Mehdi Azabou,
  Eva~L. Dyer, Rémi Munos, Petar Veličković, and Michal Valko.
\newblock Large-scale representation learning on graphs via bootstrapping.
\newblock {\em arXiv preprint arXiv: Arxiv-2102.06514}, 2021.

\bibitem[\protect\citeauthoryear{Velickovic \bgroup \em et al.\egroup
  }{2019}]{DGI}
Petar Velickovic, William Fedus, William~L. Hamilton, Pietro Li{\`{o}}, Yoshua
  Bengio, and R.~Devon Hjelm.
\newblock Deep graph infomax.
\newblock In {\em {ICLR} (Poster)}. OpenReview.net, 2019.

\bibitem[\protect\citeauthoryear{Veličković \bgroup \em et al.\egroup
  }{2017}]{GAT}
Petar Veličković, Guillem Cucurull, Arantxa Casanova, Adriana Romero, Pietro
  Liò, and Yoshua Bengio.
\newblock Graph attention networks.
\newblock {\em arXiv preprint arXiv: Arxiv-1710.10903}, 2017.

\bibitem[\protect\citeauthoryear{Wu \bgroup \em et al.\egroup }{2019}]{SGC}
Felix Wu, Amauri H.~Souza Jr., Tianyi Zhang, Christopher Fifty, Tao Yu, and
  Kilian~Q. Weinberger.
\newblock Simplifying graph convolutional networks.
\newblock In {\em {ICML}}, volume~97 of {\em Proceedings of Machine Learning
  Research}, pages 6861--6871. {PMLR}, 2019.

\bibitem[\protect\citeauthoryear{Wu \bgroup \em et al.\egroup }{2020}]{GIB}
Tailin Wu, Hongyu Ren, Pan Li, and Jure Leskovec.
\newblock Graph information bottleneck.
\newblock In {\em NeurIPS}, 2020.

\bibitem[\protect\citeauthoryear{Wu \bgroup \em et al.\egroup
  }{2021}]{survey2109}
Lirong Wu, Haitao Lin, Zhangyang Gao, Cheng Tan, and Stan.~Z. Li.
\newblock Self-supervised learning on graphs: Contrastive, generative,or
  predictive.
\newblock {\em arXiv preprint arXiv: Arxiv-2105.07342}, 2021.

\bibitem[\protect\citeauthoryear{Xu \bgroup \em et al.\egroup }{2019}]{GIN}
Keyulu Xu, Weihua Hu, Jure Leskovec, and Stefanie Jegelka.
\newblock How powerful are graph neural networks?
\newblock In {\em {ICLR}}. OpenReview.net, 2019.

\bibitem[\protect\citeauthoryear{You \bgroup \em et al.\egroup
  }{2020}]{GraphCL}
Yuning You, Tianlong Chen, Yongduo Sui, Ting Chen, Zhangyang Wang, and Yang
  Shen.
\newblock Graph contrastive learning with augmentations.
\newblock In {\em NeurIPS}, 2020.

\bibitem[\protect\citeauthoryear{Zhu \bgroup \em et al.\egroup }{2020}]{GRACE}
Yanqiao Zhu, Yichen Xu, Feng Yu, Qiang Liu, Shu Wu, and Liang Wang.
\newblock Deep graph contrastive representation learning.
\newblock {\em arXiv preprint arXiv: Arxiv-2006.04131}, 2020.

\bibitem[\protect\citeauthoryear{Zhu \bgroup \em et al.\egroup }{2021}]{GCA}
Yanqiao Zhu, Yichen Xu, Feng Yu, Qiang Liu, Shu Wu, and Liang Wang.
\newblock Graph contrastive learning with adaptive augmentation.
\newblock In {\em {WWW}}, pages 2069--2080. {ACM} / {IW3C2}, 2021.

\end{thebibliography}

\end{document}